\documentclass{article} 
\usepackage{iclr2024_conference,times}

\usepackage{amsmath,amsfonts,bm}

\def\eqref#1{equation~\ref{#1}}

\def\1{\bm{1}}

\DeclareMathAlphabet{\mathsfit}{\encodingdefault}{\sfdefault}{m}{sl}
\SetMathAlphabet{\mathsfit}{bold}{\encodingdefault}{\sfdefault}{bx}{n}

\usepackage{hyperref}
\usepackage{url}

\usepackage{boldline}

\title{Africa-Centric Self-Supervised Pre-Training for Multilingual Speech Representation in a Sub-Saharan Context}

\author{Antoine Caubrière~\& Elodie Gauthier, 
Orange Innovation, France\\
\texttt{\{antoine.caubriere,elodie.gauthier\}@orange.com} \\
}

\iclrfinalcopy 
\begin{document}

\maketitle

\begin{abstract}
We present the first self-supervised multilingual speech model trained exclusively on African speech.
The model learned from nearly 60 000 hours of unlabeled speech segments in 21 languages and dialects spoken in sub-Saharan Africa. 
On the SSA subset of the FLEURS-102 dataset, our approach based on a HuBERT$_{base}$ (0.09B) architecture shows competitive results, for ASR downstream task, compared to the w2v-bert-51 (0.6B) pre-trained model proposed in the FLEURS benchmark, while being more efficient by using 7x less data and 6x less parameters.
Furthermore, in the context of a LID downstream task, our approach outperforms FLEURS baselines accuracy by over 22\%.

\end{abstract}

\section{Introduction}

Popular self-supervised learning (SSL) approaches have shown their potential to handle multilingual speech recognition (ASR) and are capable of achieving top performance (\citet{xlsr53}; \citet{chung2021w2vbert}; \citet{pratap2023scaling}).
They enable a model to be pre-trained on a vast amount of unlabeled data, producing richer audio representation for training downstream models, compared to standard features such as MFCCs or filterbanks.
A pre-trained model can be used as a speech encoder with a fine-tuning or as a feature extractor by freezing its weights during the downstream task training. In any case, the performance of the downstream task models will be affected by the characteristics of the speech data used for pre-training \citep{zhao2022LRLASRSSL}.

Although  \cite{pires2019multilingualmbert} already demonstrated, five years ago, that transfer learning from resource-rich to resource-poor languages is more effective when the languages share similar typological features and, later, \citet{joshi2020state}, revealed that 48\% of the typological features indexed in the World Atlas of Language Structures (WALS) classification project\footnote{\url{https://wals.info/feature}} do not appear in datasets, 
most of the multilingual pre-trained speech models publicly released today still are mainly learned from only very few languages, causing their over-representation at the cost of others (\citet{valk2021voxlingua107}; \citet{fleursDataset}; \citet{xls-r128}; \citet{zhang2023usm}). African languages, which have unique characteristics and are underresourced, are severely affected by this situation (\citet{Clements_Rialland_2007}; \citet{yadav2022survey}).

Fortunately, African languages gain interest in the NLP community. 
Several studies have demonstrated the effectiveness of Africa-centric pre-trained models, showing superior performance compared to large multilingual pre-trained models that are primarily trained on English (\citet{ogueji2021small}; \citet{masakhaner2}; \citet{dossou2022afrolm}; \citet{adebara2022serengeti}). 
In speech processing, several challenges and publications of new resources recently appeared (
\cite{sikasote2021bembaspeech}; \cite{tamasheq_speech}; \cite{olatunji2023afrispeech200}; \citet{wanjawa2023kenswquad}).
On the ASR downstream task, \citet{ssl_LVSR4African_2022} got better performance for several African languages when applying self-supervised techniques and multilingual modeling, compared to traditional approaches.

In line with these works, we tackle in this paper the under-representation of African languages by proposing a multilingual speech pre-trained model specifically made for performing downstream tasks in sub-Saharan Africa (SSA) languages, by only using spoken data from this region.

\section{Datasets}
\paragraph{Unlabeled} 
The pre-trained dataset we created is composed of broadcast news recordings from diverse sources publicly available on the Web, across several countries, during May 2023. 
Sometimes, the same recording could be available in different languages spoken in the country.
Data collected contained both studio recordings (controlled environment, prepared talks) and street interviews (noisy environment, spontaneous speech).
Occasionally, jingles or songs appeared in the audio content. We therefore applied a voice activity detection (VAD) tool \citep{Bredin23} to get segments containing only speech.
Finally, we gathered a dataset which comprises nearly 60\,000 hours of speech segments and covers 21 languages and variants. 
For details, see appendix \ref{dataset}.

\paragraph{Labeled}
\citet{fleursDataset} publicly released a parallel speech dataset in 102 languages and proposed it as benchmark. Data are divided in seven macro family, including a sub-Saharan Africa group. We therefore evaluate our approach on this SSA subset (FLEURS$_{SSA}$) which is composed of 20 languages, 
5 of which are present in our pre-trained dataset.

\section{Experiments}
Experiments were carried out using the well-known HuBERT approach \citep{hubert} with the base configuration (90M parameters).
The pretraining task was achieved using the unlabeled data and the fairseq toolkit \citep{DBLP:journals/corr/abs-1904-01038} through two successive iterations on 4 A100 40Gb GPUs.
The first iteration was trained for 275k steps, using a K-means clustering computed on the MFCCs extracted from the training set as target labels. 
The second iteration was trained for 500k steps, 
and used embeddings from the 6th transformer layer using 600 hours of the training set.
The ratio between languages has been preserved.
The finally obtained pre-trained model is publicly available\footnote{\url{https://huggingface.co/Orange/}}.

For downstream task training, we used the SpeechBrain toolkit \citep{ravanelli2021speechbrain}. 
The final pretrained model is considered as a speech encoder and is fully fine-tuned with two 1024 linear layers and a softmax output at the top.
A first pool of speech recognition system (60k$_{(0.09B)}$) is obtained by a direct fine-tuning of the whole model on each language of the FLEURS dataset.
A second pool (60k$_{FT-ALL (0.09B)}$) is then obtained by first jointly fine-tuning on all languages before fine-tuning again on each language.

Following the methodology of the FLEURS paper \citep{fleursDataset} and to be consistent with their results, we did not rescore the hypothesis with a language model.
Average character error rates (CERs) obtained on the 20 languages of the FLEURS$_{SSA}$ test set are given in  table~\ref{tab-res-small}. The detailed scores per language are provided in appendix \ref{results}.

\begin{table}[h]
\centering
\begin{tabular}{lccc|cc}
\hline
 & \multicolumn{3}{c}{CER} & \multicolumn{2}{c}{WER}  \\
 & \multicolumn{1}{c}{60k$_{(0.09B)}$} & \multicolumn{1}{c}{60k$_{FT-ALL (0.09B)}$} & \multicolumn{1}{c}{FLEURS$_{w2v-bert (0.6B)}$} & \multicolumn{1}{c}{60k} & \multicolumn{1}{c}{60k$_{FT-ALL}$} \\ \hline \hline
 \textit{average} & 15.8 & 13.8 & 13.6 & 52.3 & 47.7\\
  \hline
    \end{tabular}
  \caption{\label{tab-res-small}Average results on SSA subpart of FLEURS-102 test set. \textit{(detailed results in appendix)}}
\end{table}


Results show that a model that is six times smaller and trained with seven times less data can achieve a performance level that is very close to the best baseline of FLEURS. 
This model is a step in the direction of more specific but cost-effective pre-trained approaches.


To ensure the quality of the speech representation, we fine-tuned our pretrained model using SpeechBrain for a language identification (LID) downstream task. 
We employed adaptive average pooling to produce output with shape \textit{[Batch,1,20]} and we applied a softmax. We call this model 60K$_{LID}$.
The model is trained for 15 epochs on the 20 languages of the FLEURS$_{SSA}$ subset. 

We also propose a second scenario where we employed adaptive average pooling to produce output with shape \textit{[Batch,1,768]}, with the addition of two linear layers to smoothly decrease the dimension from 768 to 256 then from 256 to 20. We call this model 60K$_{LID-smooth}$.
It is trained under the same conditions as 60K$_{LID}$.
Accuracy for both scenarios is presented in Table~\ref{tab-res-LID}.

\begin{table}[h]
\centering
\begin{tabular}{lcccc}
 \hline
     & FLEURS$_{w2v-bert}$ & FLEURS$_{mSLAM}$ & 60K$_{LID}$ & 60K$_{LID-smooth}$ \\
 \hline
 \textit{FLEURS$_{SSA}$} & 59.1 & 62.2 & 84.9  & 90.4 \\
  \hline
    \end{tabular}
  \caption{\label{tab-res-LID}LID accuracy on SSA subset of FLEURS-102 test set.}
\end{table}

Experiments have shown that our pre-trained model yields significantly improved results.
This improvement can be attributed to the model's specialization in SSA languages. 
Specifically, we utilized only SSA speech data for pretraining and, during fine-tuning, the model was trained solely on the 20 SSA languages from the FLEURS dataset, rather than the full dataset of 102 languages.

The results obtained on both downstream tasks suggest that our models produce relevant multilingual speech representations within the specific context of SSA languages.

\section{Conclusion}


To the best of our knowledge, we present the first open source SSL model exclusively pre-trained on sub-Saharan African languages. By only focusing on African speech that contains specific features unobserved in other languages spoken in the world, we improved the robustness on the ASR downstream task for SSA languages. 
While we obtain similar results on the overall SSA subset than the best model presented in the FLEURS paper (w2v-BERT-51), yet our approach is more efficient by using much less data and a reduced number of parameters for pre-training.
On a LID downstream task, results show that our specialized model trained on the SSA context performs better than the two FLEURS baselines, by obtaining more than 22\% in absolute accuracy.


\bibliography{iclr2024_conference}
\bibliographystyle{iclr2024_conference}

\newpage
\appendix
\section{Pre-trained dataset detailed \label{dataset}}

In the following table  \ref{tab-lang}, we present the languages distribution in the pre-training set. \\
We applied automatic segmentation 
on the raw recordings. \\
For the French language set, only African accented French was used. \\
"Unknown" row at the end of the table means speech recordings with language mixing. \\
No automatic LID has been applied to the segments.

\begin{table}[!h]
\centering
\begin{tabular}{|l|c|r|}
\hline
\textbf{Language} & \textbf{ISO-3} & \textbf{Hours} \\ 

\hline
 Bambara & bam & 2\,552 \\
 Dyula & dyu & 14 \\
 French & fra & 5\,670 \\
 Fula & ful & 702 \\
 Fulfulde & ffm & 727 \\
 Fulfulde & fuh & 446 \\
 Gulmancema & gux & 13 \\
 Hausa & hau & 9\,211 \\
 Kinyarwanda & kin & 8\,046 \\
 Kituba & ktu & 647 \\  
 Lingala & lin & 1\,269 \\
 Luba-Lulua & lua & 675 \\
 Mossi & mos & 13 \\
 Maninkakan & mwk & 791 \\
 Sango & sag & 1\,268 \\
 Songhai & son & 780 \\
 Swahili & swc & 706 \\
 Swahili & swh & 13\,926 \\
 Tamasheq & taq & 1\,212 \\
 Wolof & wol & 64 \\
 Zarma & dje & 567 \\
 \textit{Unknown} & --- & 10\,272 \\
 \hline \textit{Total} & --- & \textit{59\,572} \\
\hline
\end{tabular}
\caption{\label{tab-lang}
Languages distribution in the pre-training set. 
}
\end{table}


\newpage
\section{Detailed results on SSA subpart of Fleurs-102\label{results}}
Results listed below are obtained when applying monolingual fine-tuning on each sub-Saharan African languages provided in the Test set of FLEURS benchmark. \\
Scores in bold show the best result depending on the approach.
We show character error rate (CER) scores along with word error rates (WERs).

\begin{table}[h]
\centering
\begin{tabular}{lrr|rr}
\hline
 & \multicolumn{2}{c}{CER} & \multicolumn{2}{c}{WER$^*$}  \\
Language & \multicolumn{1}{c}{60k$_{(0.09B)}$} & \multicolumn{1}{c}{60k$_{FT-ALL (0.09B)}$} & \multicolumn{1}{c}{60k} & \multicolumn{1}{c}{60k$_{FT-ALL}$} \\ \hline \hline
\multicolumn{5}{l}{\textit{Seen languages}} \\
\hline
Fula          & 21.2 & \textbf{17.8} & 61.9 & \textbf{56.4} \\ 
Hausa         & 10.5 & \textbf{9.0}  & 32.5 & \textbf{29.4} \\ 
Lingala       &  8.7 & \textbf{6.9}  & 24.7 & \textbf{20.9} \\ 
Swahili       &  7.1 & \textbf{5.5}  & 23.8 & \textbf{20.3} \\ 
Wolof         & 19.4 & \textbf{17.0} & 55.0 & \textbf{50.7} \\ \hline \hline 
\textit{average} & 13.4 & \textbf{11.2} & 39.6 & \textbf{35.5} \\ \hline 
\hline
\multicolumn{5}{l}{\textit{Unseen languages}} \\
\hline
Afrikaans     & 23.3 & \textbf{20.3} & 68.4 & \textbf{62.6} \\
Amharic       & 15.9 & \textbf{14.9} & 52.7 & \textbf{49.0} \\ 
Ganda         & 11.5 & \textbf{10.7} & 52.8 & \textbf{50.3} \\ 
Igbo          & 19.7 & \textbf{17.2} & 57.5 & \textbf{52.9} \\ 
Kamba         & 16.1 & \textbf{15.6} & 53.9 & \textbf{53.7} \\ 
Luo           &  9.9 &  \textbf{8.2} & 38.9 & \textbf{34.9} \\ 
Northen-Sotho & 13.5 & \textbf{11.7} & 43.2 & \textbf{38.9} \\ 
Nyanja        & 13.3 & \textbf{10.9} & 54.2 & \textbf{48.3} \\ 
Oromo         & 22.8 & \textbf{20.1} & 78.1 & \textbf{74.8} \\ 
Shona         & 11.6 &  \textbf{8.3} & 50.2 & \textbf{39.3} \\ 
Somali        & 21.6 & \textbf{19.7} & 64.9 & \textbf{60.3} \\ 
Umbundu       & 21.7 & \textbf{18.8} & 61.7 & \textbf{54.2} \\ 
Xhosa         & 11.9 &  \textbf{9.9} & 51.6 & \textbf{45.9} \\ 
Yoruba        & 24.3 & \textbf{23.5} & 67.5 & \textbf{65.7} \\ 
Zulu          & 12.2 &  \textbf{9.6} & 53.4 & \textbf{44.9} \\ \hline \hline 
\textit{average} & 16.6 & \textbf{14.6} & 56.6 & \textbf{51.7} \\ \clineB{1-5}{2} \vspace{-5pt} \\ 
\textit{overall average} & 15.8 & \textbf{13.8} & 52.3 & \textbf{47.7} \\ \hline 
  
\end{tabular}
\caption{\label{tab-res-full}Results obtained on the Test set of the 20 languages from the SSA subpart of FLEURS-102. 
}
\end{table}


\end{document}